\pdfoutput=1

\documentclass[11pt]{article}

\usepackage{acl}

\usepackage{times}
\usepackage{latexsym}
\usepackage{subcaption}
\usepackage{tikz}
\usepackage{pgfplots}
\pgfplotsset{compat=1.18}
\usepgfplotslibrary{statistics}
\usepgfplotslibrary{groupplots}
\usepackage{enumitem}
\usepackage[normalem]{ulem}

\usepackage{amsmath}

\usepackage{amsmath,amsfonts,bm}

\def\eqref#1{equation~\ref{#1}}

\def\1{\bm{1}}

\def\vq{{\bm{q}}}

\def\mA{{\bm{A}}}

\def\mK{{\bm{K}}}

\def\mM{{\bm{M}}}

\def\mQ{{\bm{Q}}}

\def\mV{{\bm{V}}}

\DeclareMathAlphabet{\mathsfit}{\encodingdefault}{\sfdefault}{m}{sl}
\SetMathAlphabet{\mathsfit}{bold}{\encodingdefault}{\sfdefault}{bx}{n}

\usepackage[T1]{fontenc}

\usepackage[utf8]{inputenc}

\usepackage{microtype}

\usepackage{inconsolata}

\usepackage{graphicx}

\title{Simultaneous Masking, Not Prompting Optimization: \\
\vspace{2pt}
A Paradigm Shift in Fine-tuning LLMs for Simultaneous Translation}

\author{ Matthew Raffel \: \: Victor Agostinelli \: \: Lizhong Chen \\ 
Oregon State University \\
\texttt{\{raffelm, agostinv, chenliz\}@oregonstate.edu}
}

\begin{document}
\maketitle
\begin{abstract}
Large language models (LLMs) have achieved state-of-the-art performance in various language processing tasks, motivating their adoption in simultaneous translation.  Current fine-tuning methods to adapt LLMs for simultaneous translation focus on prompting optimization strategies using either data augmentation or prompt structure modifications. However, these methods suffer from several issues, such as unnecessarily expanded training sets, computational inefficiency from dumping the key and value cache, increased prompt sizes, or restriction to a single decision policy. To eliminate these issues, in this work, we propose \textit{SimulMask}, a new paradigm for fine-tuning LLMs for simultaneous translation. It utilizes a novel attention mask approach that models simultaneous translation during fine-tuning by masking attention for a desired decision policy.  Applying the proposed SimulMask on a Falcon LLM for the IWSLT 2017 dataset, we have observed a significant translation quality improvement compared to state-of-the-art prompting optimization strategies on five language pairs while reducing the computational cost.
\end{abstract}

\section{Introduction}

Simultaneous translation refers to the process of producing a target output translation concurrently with an oncoming source input. In our increasingly interconnected world, where communication across languages in real-time is desired, simultaneous translation is becoming a requirement.  
As such, there is a need for machine learning models to fill this role.  

Current literature has primarily focused on adapting end-to-end Transformer models \cite{vaswani-et-al-2017} to overcome the difficulties of simultaneous machine translation (SimulMT) due to their reduced parameter counts and greater inference speed\cite{ma2020simulmt}. However, the recent successes of large language models (LLMs) \cite{touvron2023llama, jiang2023mistral, almazrouei2023falcon} has prompted preliminary research applying them to SimulMT through fine-tuning and inference techniques \cite{agostinelli2023simul, wang2023simultaneous, koshkin2024transllama, wang2024conversational, guo2024sillm}. Unfortunately, most modern works have neglected the computational increases created by dumping the target sequence's key and value (KV) cache \cite{wang2024conversational}.  Furthermore, there has yet to be a universal approach to fine-tuning LLMs for SimulMT that is not unnecessarily computationally expensive by either expanding the dataset through data augmentation, a process referred to as prefix fine-tuning \cite{agostinelli2023simul, wang2023simultaneous, koshkin2024transllama} or increasing the prompt length through prompt restructuring \cite{koshkin2024transllama, wang2024conversational}.  

The lack of an efficient fine-tuning strategy of LLMs for SimulMT has led us to propose a new paradigm, referred to as \textit{SimulMask}.  SimulMask is a novel attention mask to model SimulMT during fine-tuning by redistributing the attention under a decision policy. By design, SimulMask is broadly applicable to both flexible and fixed decision policies, creating a path forward for future work to build upon it.  Furthermore, if we avoid injecting positional information into the keys and values through a modified ALiBi \cite{press2021train}, SimulMask allows for KV caching during SimulMT without accuracy degradation.

To validate the efficacy of SimulMask, we fine-tuned and evaluated 1.3 billion parameter Falcon models pre-trained on the RefinedWeb dataset using SimulMask \cite{almazrouei2023falcon, refinedweb} and compared them against identical Falcon models that adopt existing prefix fine-tuning or prompt restructuring methods on the IWSLT 2017 dataset \cite{cettolo-etal-2017-overview}.  From the results, we demonstrate models fine-tuned with SimulMask outperform prefix fine-tuning and prompt restructuring models at SimulMT for a given latency regime with a reduced computational cost.

The main contributions of the paper include:
\begin{enumerate}[topsep=1pt, itemsep=-1ex]
    \item Providing insights on the shortcomings of current methods in adapting LLMs to SimulMT.
    \item Proposing a novel attention masking approach to fine-tune SimulMT LLMs that enables efficient training and inference.
    \item Demonstrating the efficacy of our approach in terms of translation quality and computational costs by evaluating them on multiple language pairs across varied latencies.
\end{enumerate}

\section{Background and Related Work}
\subsection{Masked Transformer Self-Attention}
We briefly review self-attention functionality in Transformers \cite{vaswani-et-al-2017} focusing on masking behavior in Equation \ref{eq:attn}. $\mM$ is defined as an optional attention mask. 

\begin{equation}
    \mA = \text{softmax}\left(\frac{\mQ \mK^T+\mM}{\sqrt{d_{head}}}\right)\mV
    \label{eq:attn}
\end{equation}

Critically, $\mM$ functions by modeling context limitations or time-based dependencies that might exist during inference but do not exist during training/fine-tuning. For generative Transformer blocks or architectures, $\mM$ is defined as a causal attention mask, where each entry, $\mM_{ij}$, is represented by Equation \ref{eq:causalattn} to avoid attending to the future. 
\begin{equation}
    \mM_{ij} = \begin{cases}
        0, &\text{if } j \leq i \\
        -\infty,                          &\text{otherwise}
    \end{cases}
    \label{eq:causalattn}
\end{equation}
\subsection{Simultaneous Translation}
SimulMT is dictated by read-write decision policies, whereby a model will wait a specific amount of time before alternating between reading and writing in fixed or flexible intervals. One fixed decision policy for SimulMT that is broadly adopted as a common baseline to build on due to its effectiveness and simplicity is the wait-k policy \cite{ma-etal-2019-stacl}.  As the name suggests, the wait-k policy will wait for k words before alternating between writing and reading a word.  Although effective in SimulMT, alternative adaptive policies have gained traction, which base reading and writing on an auxiliary model or a predefined set of rules \cite{cho2016can, gu-etal-2017-learning, zheng2019simpler}. While capable of impressive results, such adaptive policies often incur additional computational costs.

A Transformer is trained for a SimulMT decision policy by masking attention scores in the encoder self-attention and the decoder cross-attention.  In the case of the encoder self-attention, each source \textit{token} is prevented from attending to future source tokens following a decision policy  \cite{ma-etal-2019-stacl}.  An example mask is provided in Appendix \ref{app:enc}.

Alternatively, in decoder cross-attention, each target token is prevented from attending to future source \textit{hidden states} following the decision policy \cite{papi-etal-2022-simultaneous}.  Equation \ref{eq:decattn} expresses each entry of the decoder cross-attention mask, $\mM_{tj}$.
\begin{equation}
    \mM_{tj} = \begin{cases}
        0, &\text{if } j \leq f(t) \\
        -\infty,                          &\text{otherwise}
    \end{cases}
    \label{eq:decattn}
\end{equation}
In Equation \ref{eq:decattn}, $f(t)$, is a decision policy function that denotes the cumulative number of source hidden states to read when predicting target token $t$.  

\subsection{Applying LLMs to SimulMT}
LLMs have demonstrated remarkable performance on neural machine translation  (NMT) \cite{moslem-etal-2023-adaptive, vilar-etal-2023-prompting, xu2023paradigm, zhang-etal-2023-machine, iyer-etal-2023-towards}. Such successes have prompted recent works to extend the reach of LLMs into the realm of SimulMT \cite{agostinelli2023simul, wang2023simultaneous, koshkin2024transllama, wang2024conversational, guo2024sillm}. LLMs are especially promising for the field of SimulMT due to their strong understanding of language semantics and meaning. Intuitively, SimulMT LLMs inject holistic linguistic knowledge that could allow for correct translation decisions when facing difficult contextual obstacles (e.g., translating a verb in a target language without access to that verb in the source language).  

Unfortunately, Equation \ref{eq:decattn} is no longer effective in modeling SimulMT for decoder-only LLMs as with the decoder of a classical Transformer. The reason is that Equation \ref{eq:decattn} is constructed specifically for the cross-attention calculation between keys exclusively from the source and queries exclusively from the target, as in Transformers.  In contrast, LLMs perform self-attentions involving the prompt, the source, and the target concurrently. Equation \ref{eq:decattn} can no longer properly mask the source (keys) from the target (queries), due to the additional prompt and target sequences in the keys and the additional prompt and source sequences in the queries.  Furthermore, Equation \ref{eq:decattn} does not enforce the autoregressive language modeling behavior of LLMs. As such, alternative means to model SimulMT have been proposed, leveraging prompting optimization.

\section{Prompting Optimization Methods}
Current methods of fine-tuning LLMs for SimulMT fall under prompting optimization. We define prompting optimization as either employing data augmentation to help with prompting \cite{koshkin2024transllama, wang2023simultaneous, agostinelli2023simul} or redefining the prompt structure \cite{wang2024conversational, koshkin2024transllama} to somewhat simulate SimulMT. 

\subsection{Data Augmentation}
Prompting optimization focusing on data augmentation resorts to subdividing each sentence in a dataset into multiple partial sentence pairs. These partial sentence pairs mimic SimulMT, as SimulMT produces outputs with a partial input.  We label such a method as \textit{prefix fine-tuning}, and although the high-level procedure is identical amongst current works, the algorithms employed to obtain these partial sentence pairs are unique.  In the case of \citet{agostinelli2023simul}, each source-target sentence pair is subdivided according to the wait-k policy such that if we order the new samples from smallest to largest, each subsequent sentence pair will have one additional target word and source word so long as the end of the target or source is not reached.   Upon completion there will be $\texttt{max}(|S|-(k-1), |T|)$ sentence pairs, where $|S|$ and $|T|$ are the original source and target sequence lengths. The approach requires the model to predict only the final target word in the sequence during fine-tuning.

Alternatively, \citet{wang2023simultaneous} randomly sampled a subset of sentence pairs from the dataset and truncated the source sentence to be 20\% to 80\% of the full length according to a uniform distribution.  They obtained the respective target translations by prompting ChatGPT (gpt-3.5-turbo).  The new truncated source-target sentence pairs were then added to the complete dataset, expanding it. 

\subsection{Prompt Restructuring}
Prompting optimization that modifies the prompting structure adjusts the prompt to include the decision policy. In the case of \citet{wang2024conversational}, a conversational prompting structure is adopted for the LLM, alternating between source and target subsequences of the original complete sequences using delimiting tokens to separate regions. For instance, if we have the source sequence $S = [s_1, s_2, ..., s_n]$ and the target sequence $T=[t_1, t_2, ..., t_m]$, then one potential conversational prompt expansion could be ``$\texttt{<s>}, \texttt{[U]}, s_1, s_2, \texttt{[A]}, t_1, t_2, \texttt{</s>}, ..., \\ \texttt{<s>}, \texttt{[U]}, s_n, \texttt{[A]}, t_m, \texttt{</s>}$'', where the added \texttt{<s>, </s>, [A], [U]} are delimiting tokens. During fine-tuning, the choice of alternating subsequences is arrived at by attempting to maximize the relevant source context before each target sequence in the form of an oracle decision policy.  For instance, the prompt will ensure an arbitrary target verb prediction only after the respective source verb is read.  Some minor perturbations are added to the oracle decision policy to improve generalizability. Then, at inference, a prompt constructor provides the source sequence in fixed-size chunks. 

Similarly, \citet{koshkin2024transllama} leverages prompt restructuring; however, it also employs prefix finetuning. Like the conversational prompting structure, it constructs a fine-tuning prompt by aligning words between the source and target sequence to mimic an oracle decision policy.  However, it deviates from conversational prompting by ensuring the alignment using padding tokens in the target sequence.  Then, the causally aligned sentence prompt is subdivided using a prefix fine-tuning strategy to expand the dataset with partially filled source-target sentence pairs.  At inference, the LLM contains the decision policy outputting padding tokens whenever it requires more source context tokens.

\section{Analysis and Shortcomings of Prompting Optimization Methods}
Prompting optimization, while functional to a certain degree, is inherently deficient, possessing a host of fine-tuning and inference issues. These issues include a persistent fine-tuning-inference mismatch,  consistent positional confusion in the target sequence, and high computational costs.  

\subsection{Fine-tuning/Inference Mismatch}
A fine-tuning-inference mismatch is a mismatch between a LLM's fine-tuning and inference environments. For instance, fine-tuning a LLM for NMT where the entire sentence is available and deploying it for SimulMT where little of the sentence is available when beginning generation will create a massive inference time fine-tuning-inference mismatch. Furthermore, the LLM must be fine-tuned to accommodate KV caching, the process of caching the keys and values at inference to prevent recomputation. Overall, fine-tuning for SimulMT aims to minimize the fine-tuning-inference mismatch. 

Unfortunately, prefix fine-tuning precludes high-quality SimulMT with KV caching \cite{agostinelli2023simul, koshkin2024transllama, wang2023simultaneous} as with the continuously increasing prompt size, each key and value in the KV cache deviates more from the keys and values in its fine-tuning environment.  For example, suppose we have a LLM in the middle of SimulMT using KV caching adhering to a wait-1 policy with the following prompting structure: ``\texttt{Translate the following sentence: $s_1, s_2, ..., s_{i+1}$ [a]: $t_1, t_2, ..., t_i$}''.  Then, at the current write step, the query of $t_i$ attends to the KV cache for $\texttt{[a]:}, t_1, t_2, ..., t_{i-1}$. By construction, each key and value in the KV cache was generated in a previous time step with a different subset of the source sequence $s_1, s_2, ..., s_i$. For instance, the keys and values for delimiting token \texttt{[a]:} when it predicted $t_1$ were conditioned only on $s_1$, whereas the keys and values for $t_{i-1}$ when it predicted $t_i$ were conditioned on $s_1, s_2, ..., s_i$. However, during prefix fine-tuning, the LLM was fine-tuned to predict $t_{i+1}$ as if the KV cache for $\texttt{[a]:}, t_1, t_2, ..., t_{i-1}$ were each generated with the same subset of the source sequence $s_1, s_2, ..., s_i$.  Such fine-tuning-inference mismatch is unsolved through conventional prompting structures. 

Prompting restructuring also creates additional fine-tuning-inference mismatches.  In \citet{koshkin2024transllama} and \citet{wang2024conversational}, they all fine-tune for an oracle decision policy.  However, at inference, such an oracle decision policy is not truly achievable, creating a mismatch.  Furthermore, since the LLMs that leverage prompt restructuring encapsulate a specific oracle decision policy into their fine-tuning curriculum, extending them to alternative decision policies at inference is infeasible without incurring a mismatch. This calls for a new, flexible method adaptable to a range of decision policies while also eliminating the fine-tuning-inference mismatch. 

\subsection{Positional Confusion}
Positional confusion describes the process whereby the relative and/or global positional information during SimulMT progressively becomes incorrect.  Unfortunately, most SimulMT LLMs using KV caching suffer from this positional confusion \cite{agostinelli2023simul, koshkin2024transllama, wang2023simultaneous}. The reason is that as the source sequence grows with SimulMT, the target sequence also shifts, necessitating the target sequence's positional information to follow suit.  However, since KV caching is employed, the positional information held in the keys and values is not properly updated.  

Aligning with our previous example, for the sequence portion ``\texttt{[a]: $t_1, t_2, ..., t_i$}'', after the first prediction step, the positional distance between $s_{1}$ and \texttt{[a]:} and between $s_{1}$ and $t_1$ would be 1 and 2, respectively. Then, after the second read, where the source sequence is $s_{1}, s_{2}$, the positional distance between $s_{1}$ and \texttt{[a]:} and $t_1$ would change to 2 and 3, respectively.  However, while using KV caching, this positional distance would remain 1 and 2 in the keys and/or values for subsequent predictions, causing positional confusion.  Continuing translation would see the increased gap between the true positional distance and the stale positional distance in the KV cache.  Therefore, we need to identify an effective method to deal with positional confusion that is essential to prevent LLM hallucinations.

\subsection{Computational Inefficiency}
Avoiding KV caching and instead recomputing all the keys and values at each prediction step is the default solution for resolving the aforementioned fine-tuning-inference mismatch and positional confusion problems while employing prefix fine-tuning.  Although effective from a translation quality standpoint, doing so incurs a large computational cost, an undesirable result for streaming tasks like SimulMT, where latency is equally important.    

Outside of KV caching, the computational costs necessary for prefix fine-tuning methods are excessive. By subdividing each sample into multiple, the dataset drastically expands, contributing toward an increased cost to complete each epoch \cite{agostinelli2023simul, koshkin2024transllama, wang2023simultaneous}. Such an increase causes the duration of each epoch to rise by upwards of a factor of 10 (as exemplified in Table \ref{tab:traintime} in Section \ref{sec:compres}).  However, unlike normal methods of expanding a dataset through data augmentation, prefix fine-tuning does not actually add additional information.  It is from this added computational burden that \citet{agostinelli2023simul} and \citet{wang2023simultaneous} are forced to fine-tune with a subset of their entire prefix datasets. 

Alternatively, methods of restructuring the prompt as in \citet{koshkin2024transllama} and \citet{wang2024conversational} have computational burdens of their own.   For instance, \citet{wang2024conversational} requires adding delimiting tokens in the prompt sequence, expanding the sequence length.  Similarly, the requirement of padding tokens to induce a causal alignment between the source and target sequences, as in \citet{koshkin2024transllama}, also expands the sequence length.  Since the computational cost of the self-attention cost in the LLM scales quadratically with the sequence length, such a method is undesirable for both inference and fine-tuning.

Currently, no computationally efficient fine-tuning approach exists that enables computationally efficient inference. Identifying such a method is necessitated by the desire for low latency and high-quality translations and reducing the already high computational costs of fine-tuning LLMs.

\section{SimulMask: A Paradigm Shift}
In this work, we propose \textit{SimulMask}, which we believe could be a paradigm shift in fine-tuning LLMs for SimulMT that eschews current methods of prompting optimization. By restricting attention during fine-tuning, SimulMask efficiently solves the fine-tuning-inference mismatch and positional confusion problem.  We demonstrate SimulMask through its application for the wait-k decision policy, but it should be noted that SimulMask broadly applies to various decision policies. 

\subsection{Inference Mirrored Attention}
\label{Sec:restrictattention}
We first introduce the concept of \textit{Inference Mirrored Attention} that cleverly models SimulMT during LLM fine-tuning. Under SimulMT, the latest translation token at each prediction step is conditioned only on the running source tokens.  For conventional Transformers, specialized attention masks could achieve such conditioning; however, directly mapping these to LLMs is impossible since they fail to enforce autoregressive language modeling and cannot mask properly when the prompt, source, and target sequences are collectively included in the queries and keys. As such, prior works attempted to achieve such conditioning during fine-tuning using prompting optimization strategies littered with shortcomings.  The proposed inference mirrored attention is aimed to model SimulMT with attention masks for LLMs by mirroring the attention during inference at fine-tuning according to the chosen decision policy.  

\begin{figure}[h]
    \centering
    \begin{subfigure}[t]{0.25\columnwidth}
        \centering
        \includegraphics[width=\linewidth]{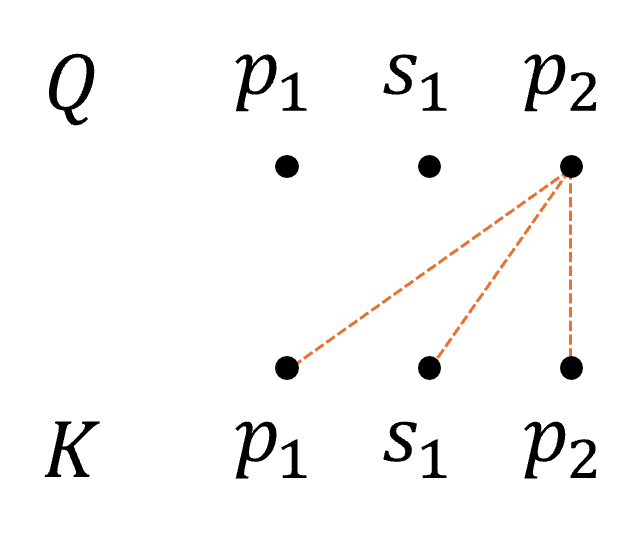}
        \caption{Inference.}
    \label{fig:inference1}
    \end{subfigure}\hfill
    \begin{subfigure}[t]{0.65\columnwidth}
        \centering 
        \includegraphics[width=\linewidth]{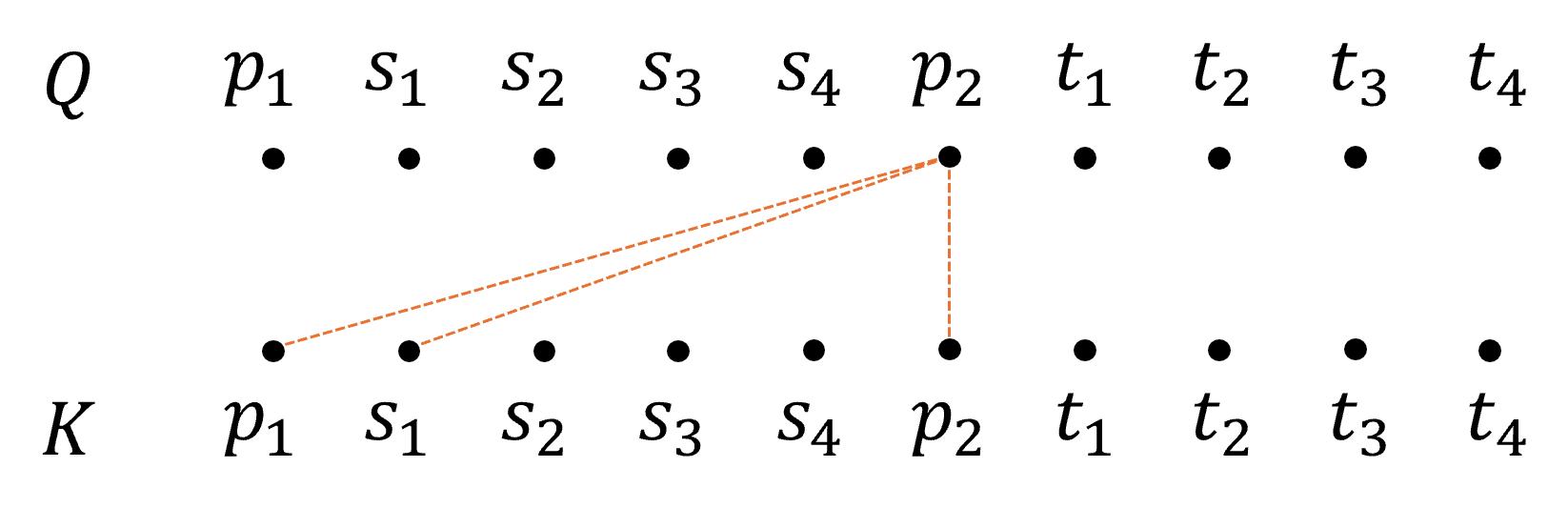}
        \caption{Fine-tuning.}
        \label{fig:finetune1}
    \end{subfigure}
\caption{Inference Mirror Attention for matching attention during inference and fine-tuning for SimulMT.}
\label{attentionalignment1}
\end{figure}

As an example, suppose we model the attention for a  wait-1 decision policy where the complete oracle input sequence is ``$p_1, s_1, s_2, s_3, s_4, p_2, t_1, t_2, t_3, t_4$''.  In the sequence, $s_1, s_2, s_3, s_4$ and $t_1, t_2, t_3, t_4$ are the 4-word source and target sequences and $p_1$ and $p_2$ are prompting regions. Then, at inference, by definition of the wait-1 policy, $p_2$ predicts $t_1$ while conditioned on the partial sequence $p_1, s_1, p_2$.  As such, as shown in Figure \ref{fig:inference1} the query of $p_2$ attends to the keys of $p_1, s_1, p_2$.  Thus, during fine-tuning, to eliminate the fine-tuning-inference mismatch, the query of $p_2$ should be limited to similarly attend to the keys of $p_1, s_1, p_2$ as shown in Figure \ref{fig:finetune1} rather than the entire source sequence. For each successive prediction step, the previously predicted target word, $t_i$, predicts the next target word, $t_{i+1}$ by conditioning on an extra source word, $s_{i+1}$, acquired from the previous read step.  To mimic such behavior at fine-tuning, the query for $t_i$ attends to identical keys as its inference step. The complete steps of this example are in Appendix \ref{app:align}.    

\subsection{SimulMask}
To achieve the above inference mirrored attention, we opt for an attention mask to restrict attention during fine-tuning to mimic an arbitrary decision policy during SimulMT. An attention mask is preferable to prompting optimization as it is flexible and directly extends the LLM causal attention mask.  We call such an attention mask SimulMask.  

\begin{figure}[h]
    \centering
    \hspace*{-1.5cm}
    \includegraphics[width=0.35\textwidth]{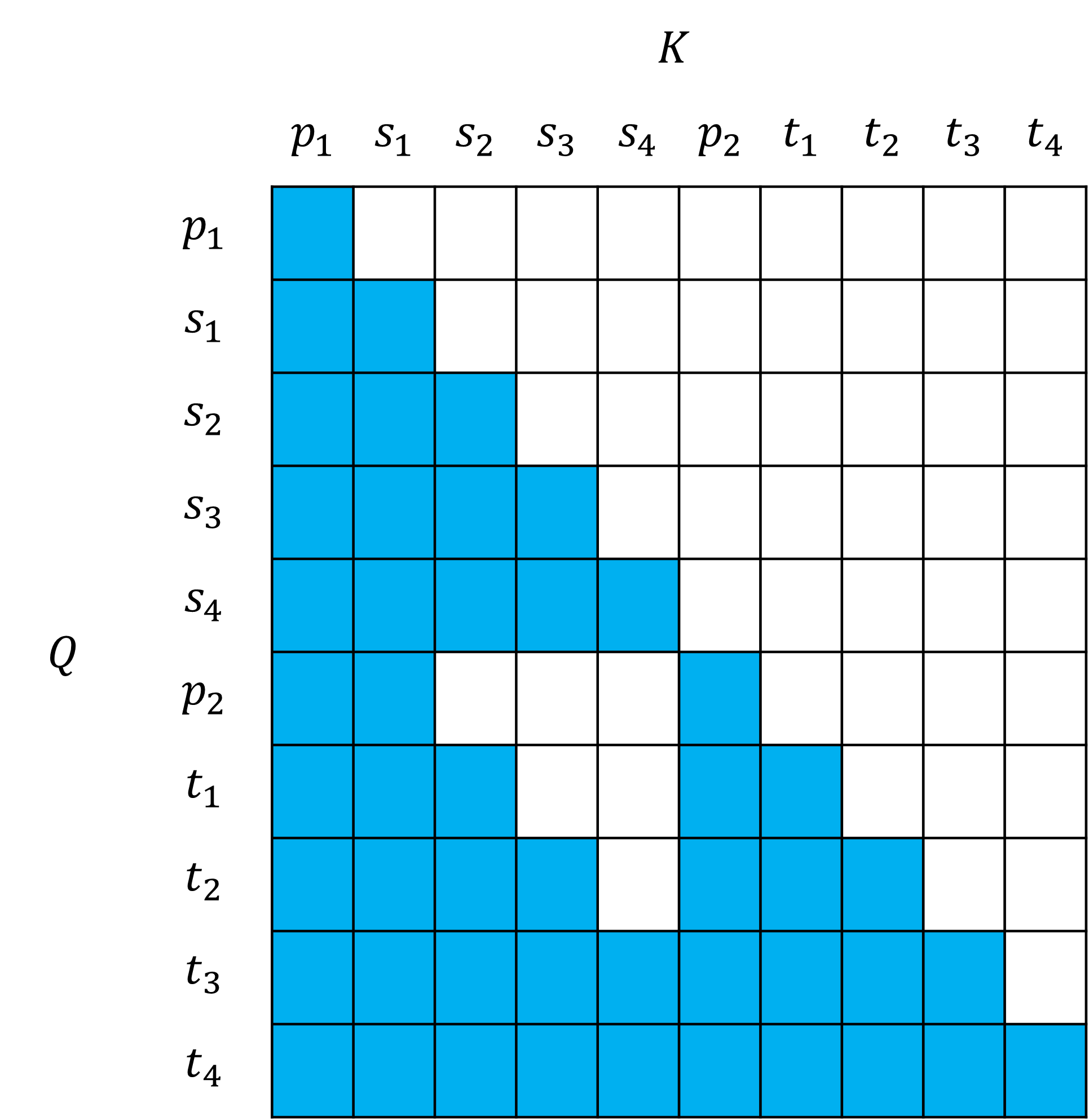}
    \caption{SimulMask for modeling SimulMT according to a wait-1 decision policy during fine-tuning.}
    \label{fig:attmat}
\end{figure}

As a demonstration, let us create a SimulMask for the wait-1 policy that extends our example from Section \ref{Sec:restrictattention}. As depicted in Figure \ref{fig:attmat}, since the LLM is autoregressive, SimulMask begins with a causal attention mask from which attention is limited to be identical to the attention during SimulMT for the source sequence.  Starting from the prompt $p_2$, from the example in Figure \ref{fig:finetune1}, $p_2$ generates the first target token, $t_1$, conditioned on $p_1, s_1, p_2$.  As such, SimulMask eliminates the attention between $p_2$ and $s_2, s_3, s_4$. Similarly, $t_1$  and $t_2$ are conditioned on $p_1, s_1, s_2, p_2, t_1$ and $p_1, s_1, s_2, s_3, p_2, t_1, t_2$, respectively.  Thus, attention is eliminated between $t_1$ and $s_3, s_4$ and $t_2$ and $s_4$.

SimulMask is a flexible scheme that supports a range of decision policies. Since each decision policy performs read/write decisions differently and each limits attention differently, this requires a unique attention mask for every sentence. However, this can be done straightforwardly. The general procedure to construct a SimulMask for a given policy and sentence consists of the following steps:
\begin{enumerate}[topsep=1pt, itemsep=-1ex]
    \item Construct a causal attention mask using Equation \ref{eq:causalattn} as a starting point for SimulMask.
    \item Starting from the intersection between the query that predicts the first target token and the first source key, apply the sub-attention mask expressed in Equation \ref{eq:decattn}.  The sub-attention mask prevents the target queries from attending to source keys following the arbitrary decision policy.
    \item Mask any non-source queries before the query predicting the first target token from attending to the source keys not included in the first read decision. Such a step is necessary to prevent the hidden states associated with these queries from holding information of the entire source sequence at later layers in the LLM.
\end{enumerate}

As reported in Section \ref{sec:compres}, the computation for constructing an arbitrary SimulMask is minor, and since SimulMask is not applied during inference, it does not impact computational cost at deployment.  Therefore, SimulMask is an efficient option for mimicking SimulMT during fine-tuning and providing low-latency translations at inference.     

\subsection{Positional Reordering}
Since positional confusion during inference is a byproduct of retaining outdated positional information in either the keys or values, bypassing it requires providing a form of positional information without injecting it directly into the sequence or KV cache.  One positioning method that satisfies such a constraint is the popular ALiBi, which supplies positional information through biases in attention \cite{press2021train}.  The bias is applied to each query-key dot product row in the attention calculation as shown in Equation \ref{eq:alibi}, where $m$ is a head-specific constant.
\begin{equation}
    \vq_i \mK^T + \mM_i + m\cdot[-(i-1),..., -1, 0]
    \label{eq:alibi}
\end{equation}
Though simple, ALiBi has demonstrated an ability to extrapolate to much larger sequence lengths than other state-of-the-art positional encodings, making it desirable for LLMs like Falcon \cite{refinedweb}, BLOOM \cite{le2023bloom}, etc.

\begin{figure}[t!]
    \centering
    \begin{subfigure}[t]{0.45\columnwidth}
        \centering
        \includegraphics[width=0.8\linewidth]{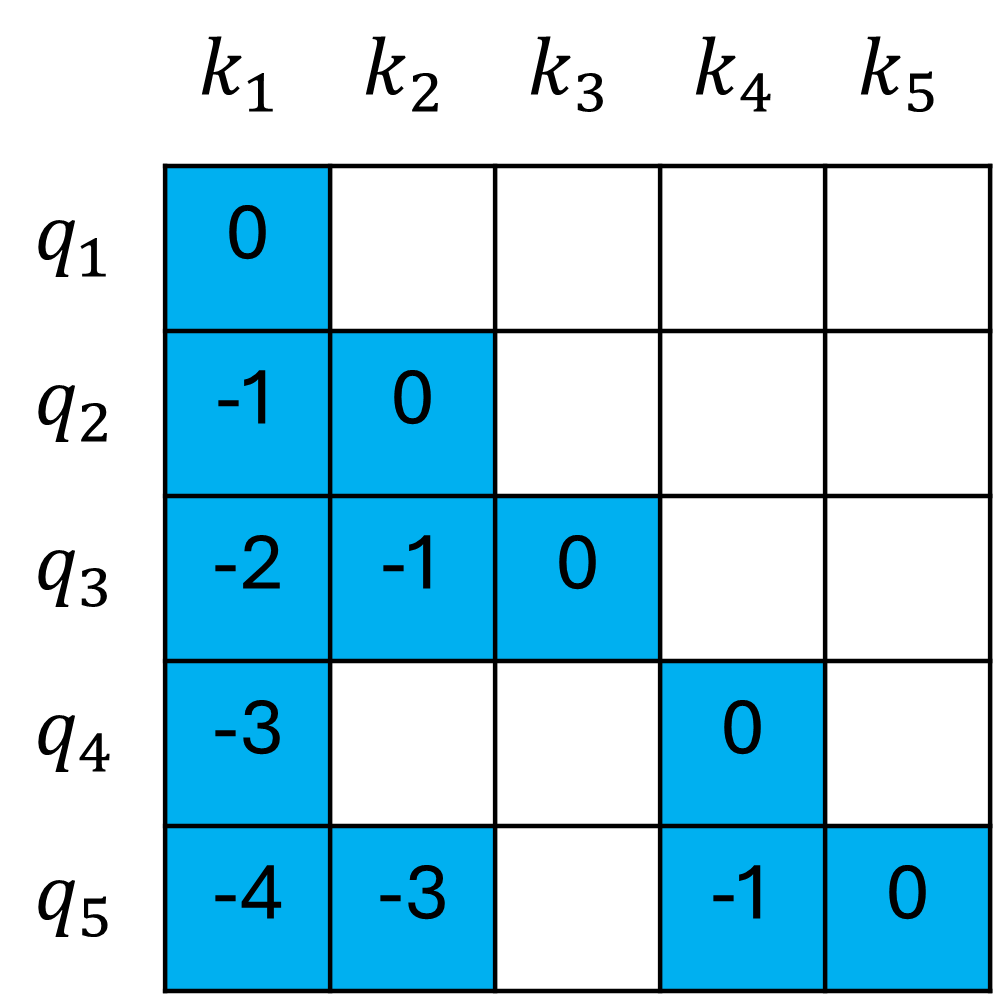}
    \caption{Original ALiBi.}
    \label{fig:alibi1}
    \end{subfigure}\quad
    \begin{subfigure}[t]{0.45\columnwidth}
           \centering 
          \includegraphics[width=0.8\linewidth]{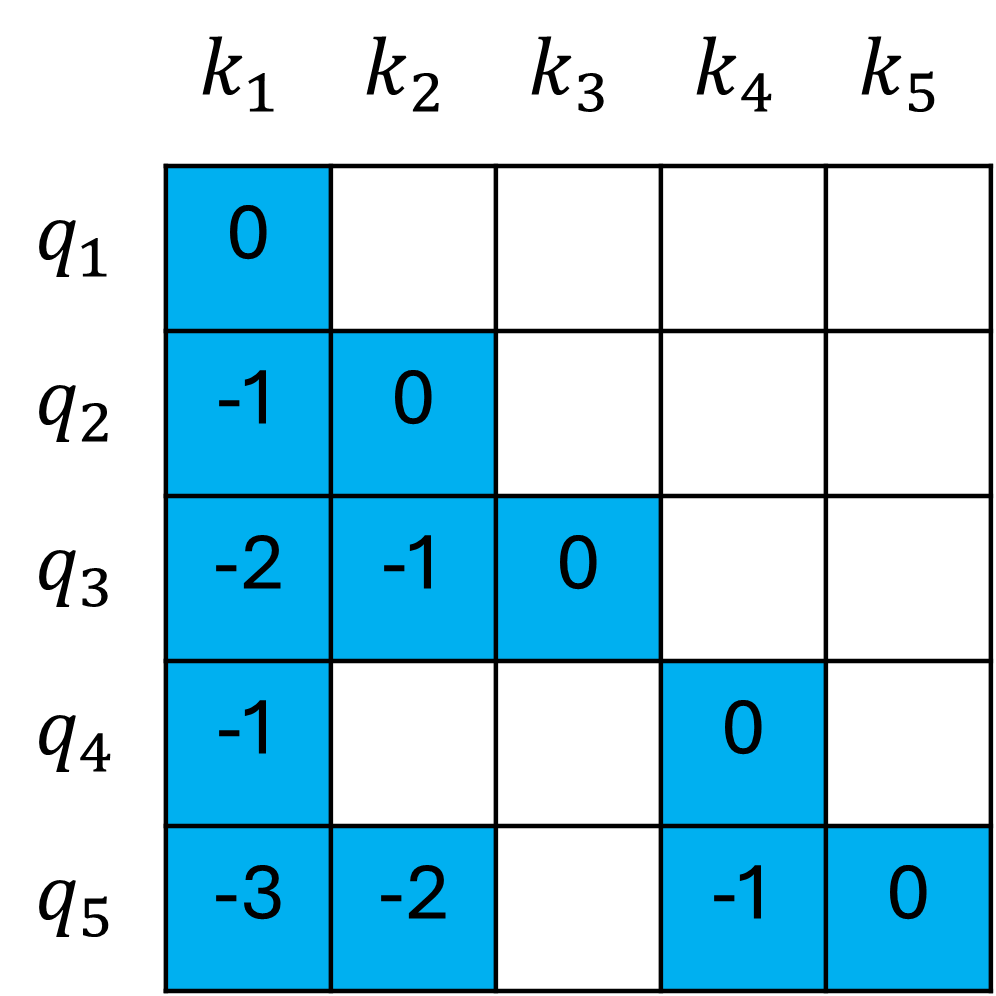}
        \caption{Modified ALiBi.}
        \label{fig:alibi2}
    \end{subfigure}
\caption{ALiBi biases with SimulMask.}
\label{alibi}
\end{figure}

Unfortunately, ALiBi, by default, does not mesh with SimulMask as SimulMask removes attention between the target queries and source keys. This removed attention creates a gap in ALiBi biases during fine-tuning that are not present at inference. An example of such a gap is provided in Figure \ref{fig:alibi1}, where both $q_4$ and $q_5$ have gaps in the position distance.    

To eliminate the bias gap, we modify ALiBi by reducing the bias values of all query rows influenced by SimulMask.  For each query row, the reduction in bias values is equivalent to the amount of attention removed along the row using SimulMask. Figure \ref{fig:alibi2} provides an example of such a modification.  In the case of $q_4$, it is no longer able to attend to $k_2$ and $k_3$; therefore, the bias on the right of the gap is reduced by 2.  Together with the modified ALiBi, SimulMask eliminates positional confusion from the LLM during SimulMT.

\section{Experimental Setup}
\begin{figure*}[t]
    \centering
    \begin{subfigure}[h]{\columnwidth}
        \centering
        \includegraphics[width=\textwidth]{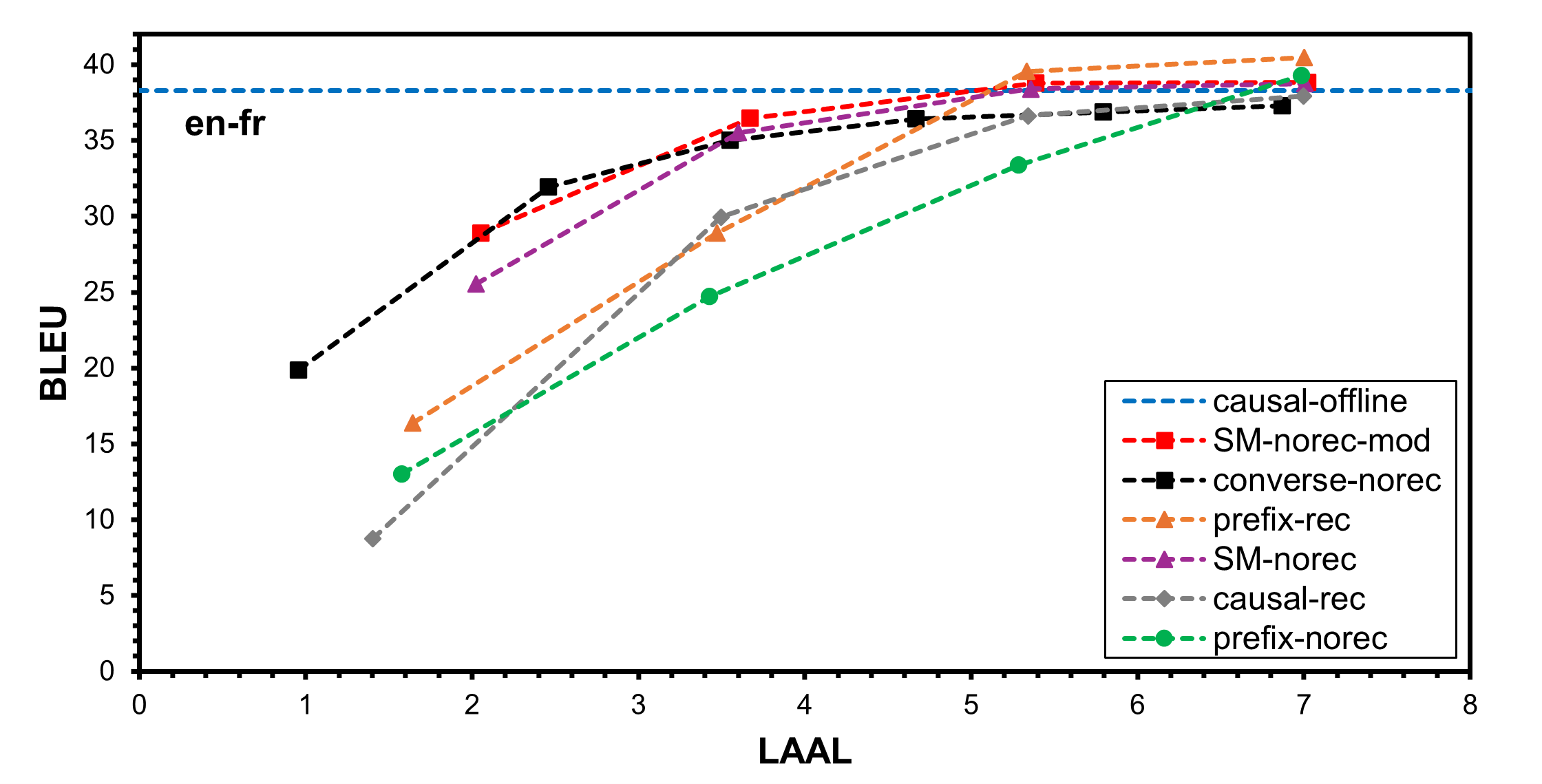}
        \caption{English-French language pair results.}
    \label{fig:en-fr}
    \end{subfigure}\hfill
    \begin{subfigure}[h]{\columnwidth}
           \centering 
            \includegraphics[width=\textwidth]{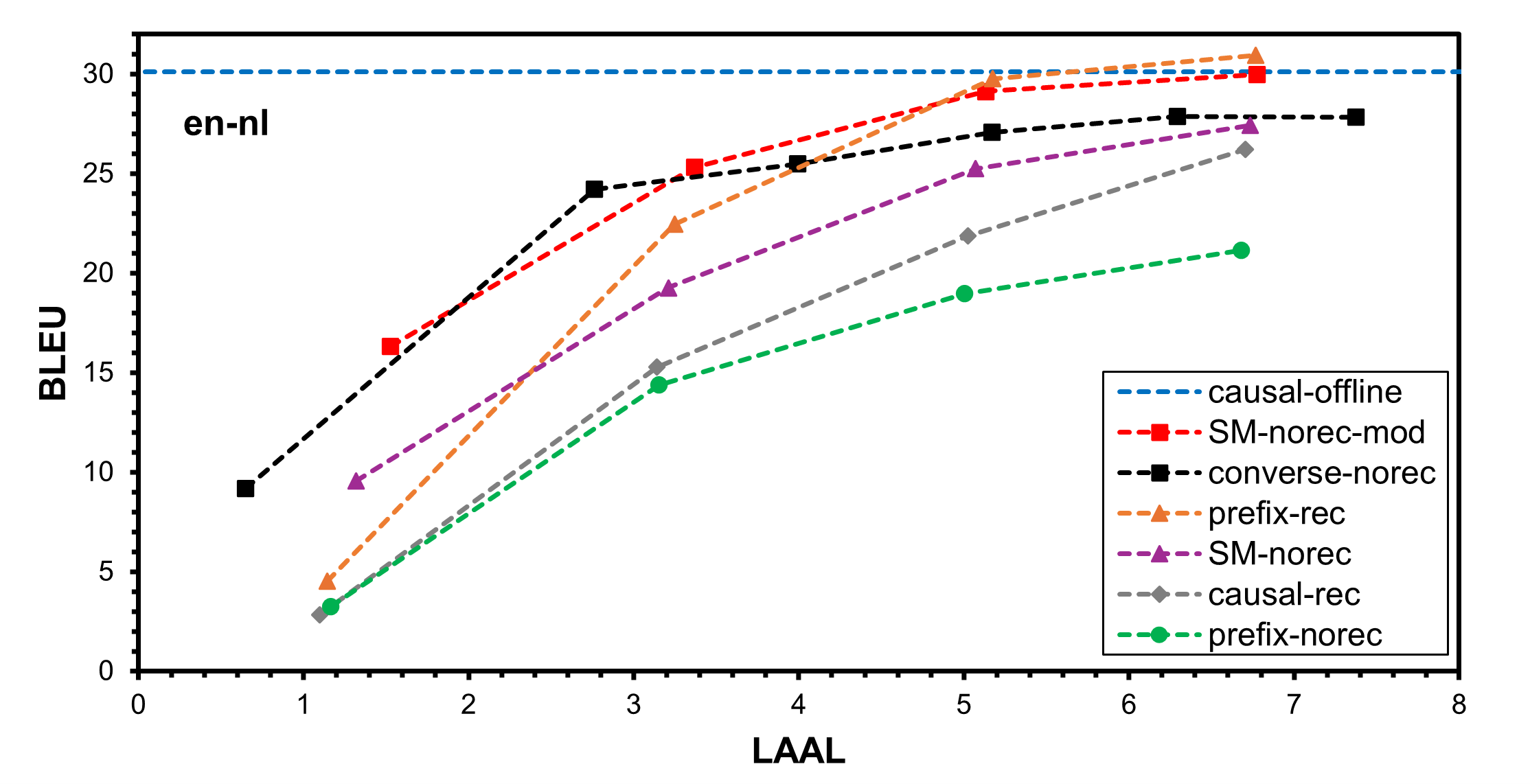}
            \caption{English-Dutch language pair results.}
        \label{fig:en-nl}
    \end{subfigure}
    \begin{subfigure}[h]{\columnwidth}
        \centering
        \includegraphics[width=\textwidth]{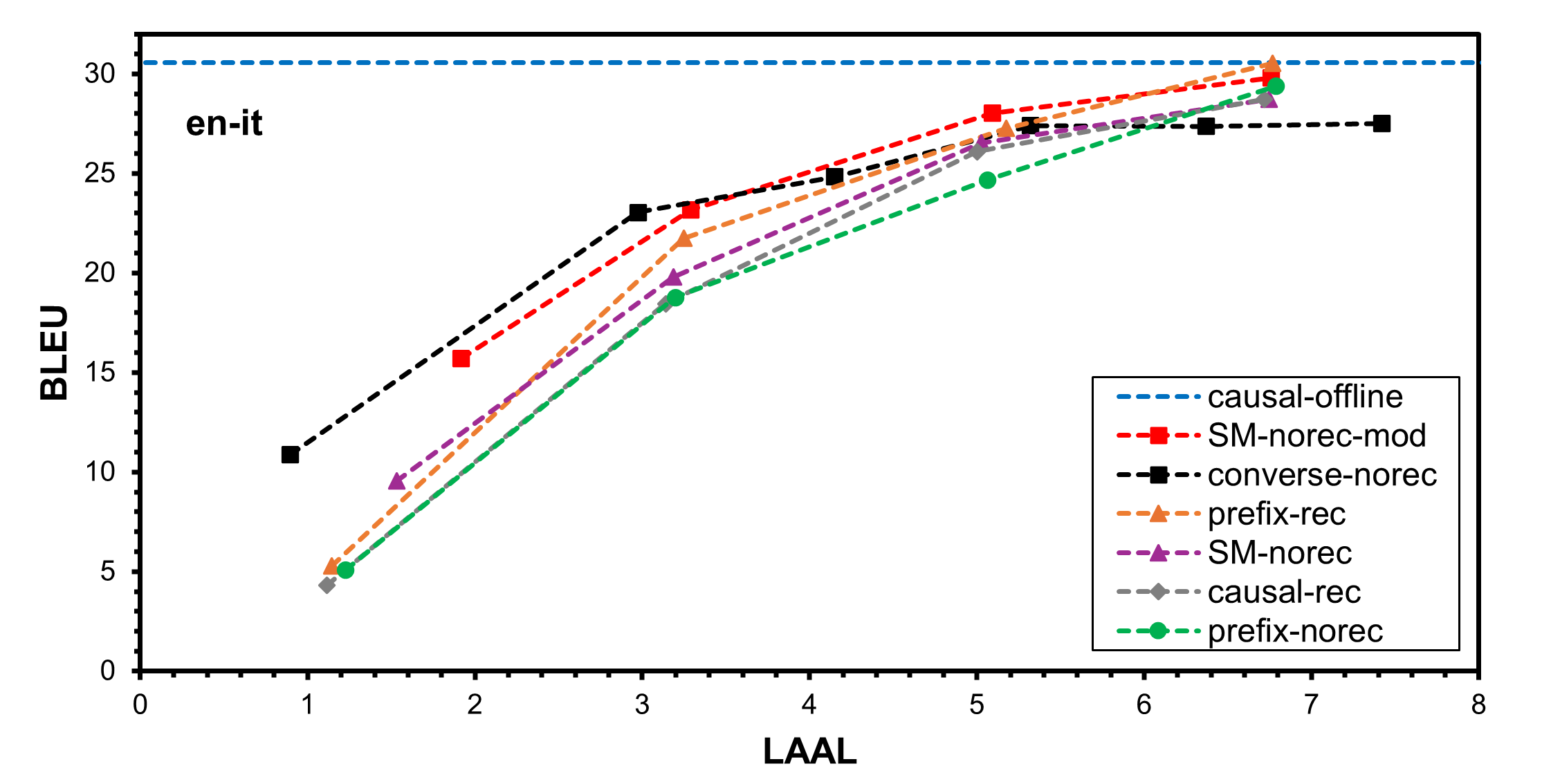}
        \caption{English-Italian language pair results.}
    \label{fig:en-it}
    \end{subfigure}\hfill
    \begin{subfigure}[h]{\columnwidth}
        \centering
        \includegraphics[width=\textwidth]{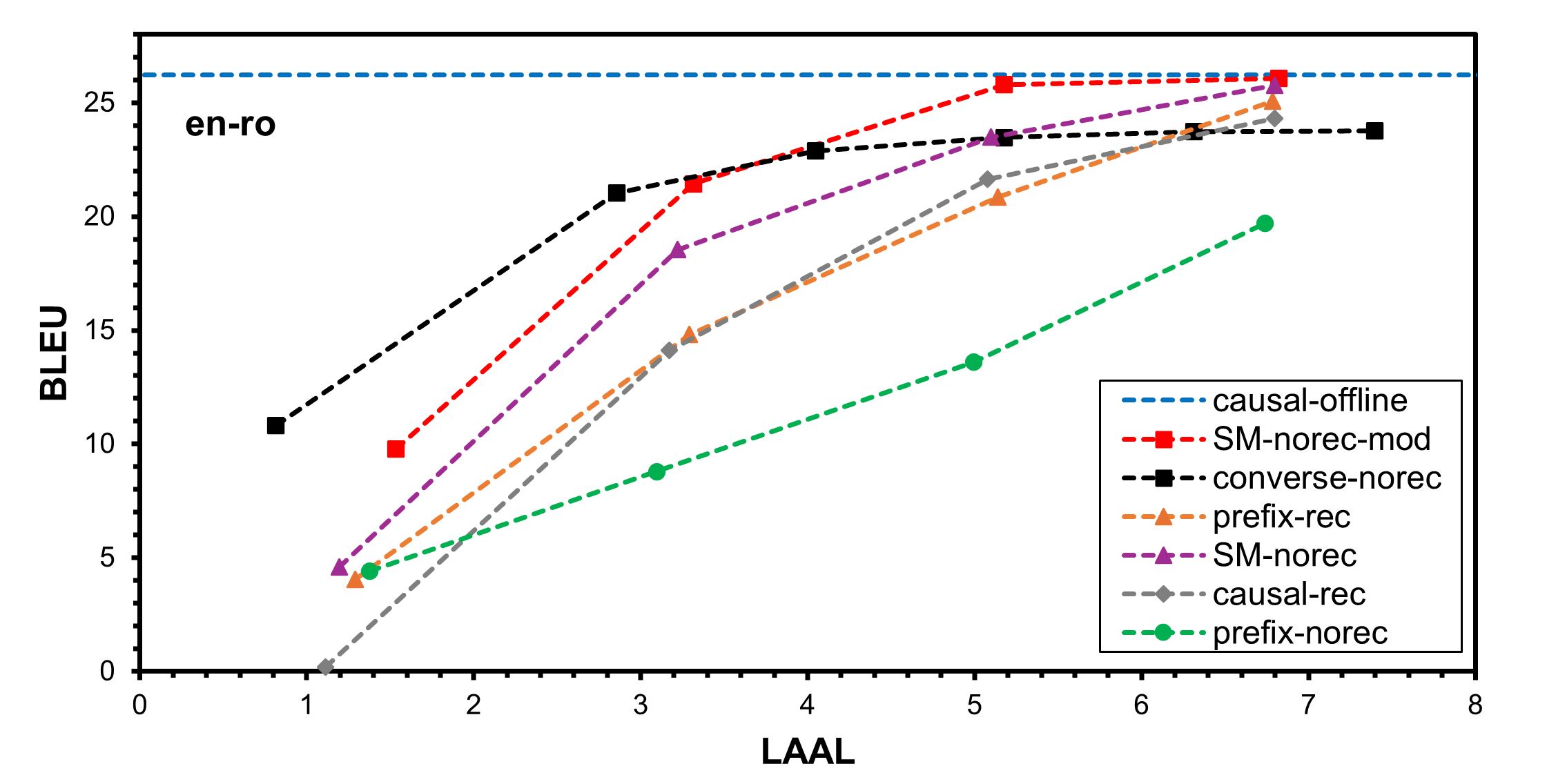}
        \caption{English-Romanian language pair results.}
    \label{fig:en-ro}
    \end{subfigure}\hfill
    \begin{subfigure}[h]{\columnwidth}
           \centering 
        \includegraphics[width=\textwidth]{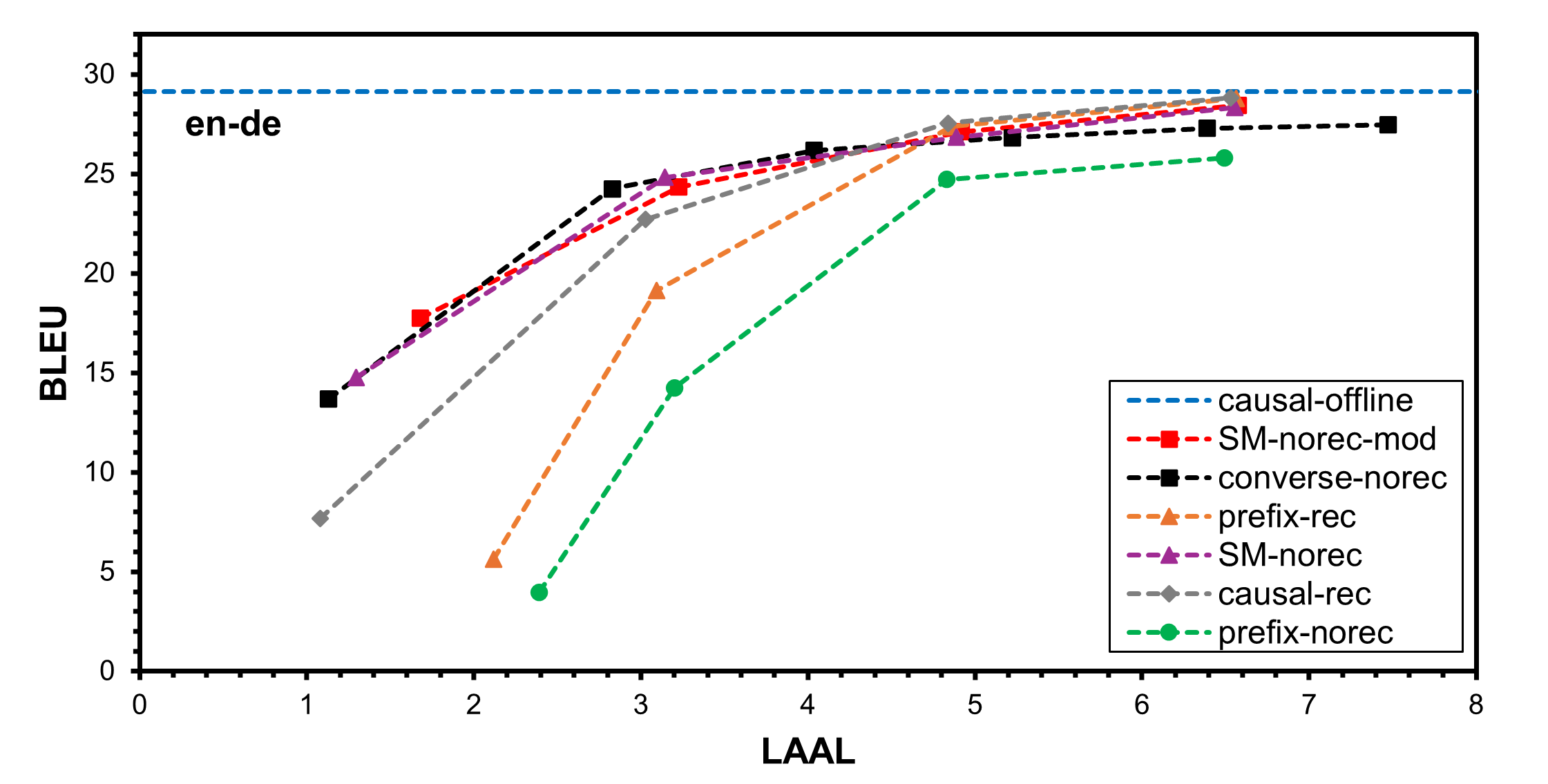}
        \caption{English-German language pair results.}
        \label{fig:en-de}
    \end{subfigure}
\caption{Translation quality plotted against latency for LLMs on the English-French, English-Dutch, English-Romanian, and English-German language pairs.}
\label{fig:translationresults}
\end{figure*}
\subsection{Fine-tuning}
Our fine-tuning was conducted with the Simul-LLM framework \cite{agostinelli2023simul}, which contains our publicly available code for SimulMask\footnotemark.  Each experiment used a 1.3 billion parameter Falcon model pre-trained on the RefinedWeb dataset \cite{refinedweb}. We compared 7 schemes:
\footnotetext{\url{https://github.com/OSU-STARLAB/Simul-LLM}}
\begin{itemize}[topsep=1pt, itemsep=-1ex]
    \item \textbf{causal-offline}: Fine-tuned with a causal attention mask and evaluated for NMT (non-SimulMT).
    \item \textbf{causal-rec}: Fined-tuned with a causal attention mask and evaluated with recomputing the KV cache.
    \item \textbf{prefix-rec} and \textbf{prefix-norec}: Fined-tuned with prefix fine-tuning and evaluated with/without recomputing the KV cache.
    \item \textbf{converse-norec}: Fined-tuned with a conversational prompting strategy and evaluated without recomputing the KV cache.
    \item \textbf{SM-norec-mod} and \textbf{SM-norec}: Fined-tuned with SimulMask with/without modifying ALiBi and evaluated without recomputing the KV cache.
\end{itemize}
 Appendix \ref{app:hyperparams} provides all model hyperparameters.
Our fine-tuning experiments included the English-French (en-fr), English-Italian (en-it), English-Dutch (en-nl), English-Romanian (en-ro), and English-German (en-de) language pairs of the IWSLT 2017 dataset \cite{cettolo-etal-2017-overview}.

\subsection{Evaluation}
We evaluated translation quality and latency for SimulMT using Simul-LLM inference agents \cite{agostinelli2023simul} interfacing with the SimulEval toolkit \cite{ma-etal-2020-simuleval}.  The translation quality was determined using detokenized BLEU with SacreBLEU \cite{post2018call} and chrF++ \cite{popovic2017chrf++}.  Latency was determined using Length-Adaptive Average Lagging (LAAL) \cite{papi-etal-2022-generation}. The computational cost of SimulMT was recorded with GFLOPs.  All metrics were obtained on a single A10 GPU with bfloat16 precision. The models fine-tuned for the wait-k policy were evaluated at a wait-k four lower, for which they were fine-tuned, as suggested by \cite{ma-etal-2019-stacl}.  

\section{Results}
\label{Sec:Results}

\subsection{Translation Quality and Latency Results}
In this section, we demonstrate the efficacy of fine-tuning with the proposed SimulMask compared with other schemes using BLEU scores and LAAL.  All wait-k model evaluations are performed across wait-\{1,3,5,7\}, and the \textit{converse-norec} is evaluated for a chunk size of 1, 3, 5, 7, 9, and 11. Figure \ref{fig:translationresults} provides the BLEU translation quality and latency results on the English-French, English-Dutch, English-Italian, English-Romanian, and English-German language pairs.  We provide the numerical BLEU and chrF++ translation quality results in Tables \ref{tab:BLEU} and \ref{tab:CHRF} of Appendix \ref{app:transresults}.

Overall, throughout Figure \ref{fig:translationresults}, the proposed \textit{SM-norec-mod} outperforms or matches the translation quality of \textit{causal-rec}, \textit{prefix-rec}, and \textit{converse-norec} across all latencies. The only major exception occurs at wait-1, where \textit{converse-norec} outperforms \textit{SM-norec-mod} on the English-Romanian language pair. This overall excellent performance in terms of translation quality underscores the importance of the proposed method.

Furthermore, Figure \ref{fig:translationresults} provides two ablation studies. The first ablation demonstrates the importance of modifying ALiBi with SimulMask for high-quality translations by comparing \textit{SM-norec-mod} with \textit{SM-norec}. For each wait-k value and language pair, \textit{SM-norec-mod} outperforms \textit{SM-norec}. Unsurprisingly, at higher wait-k values where the setting approaches NMT, the difference in BLEU scores becomes less pronounced between the models. 

A secondary ablation is provided in Figure \ref{fig:translationresults} by comparing \textit{prefix-rec} and \textit{prefix-norec}.  Doing so demonstrates that translation quality increases by recomputing the KV cache across all wait-k values.  Similarly, as with the previous ablation, the difference in the BLEU score becomes less pronounced for the higher wait-k values.   

An interesting observation is that models evaluated at lower wait-k values have their LAAL deviate from their respective k to a greater degree than those evaluated at higher wait-k.  Such an increase is a byproduct of the lower wait-k models generating longer predictions than their corresponding references.  The increased generation length is a byproduct of the model hallucinating on sequences provided insufficient contexts.    These hallucinations are most noticeable with \textit{prefix-rec} and \textit{prefix-norec} in Figure \ref{fig:en-de}.

\subsection{Compuational Saving Results}
\label{sec:compres}
Fine-tuning LLMs with SimulMask also features reduced training time compared with LLMs leveraging prefix fine-tuning or conversational prompting.  For instance, this is reflected in the fine-tuning times for one epoch on an H100 GPU on the English-French dataset of the IWSLT 2017 dataset, as reported in Table \ref{tab:traintime} \cite{cettolo-etal-2017-overview}.
\begin{table}[h!]
  \centering
  \begin{tabular}{lc}
    \hline
    \textbf{Fine-tuning Approach} & \textbf{Time (s)}\\
    \hline
    Prefix Fine-tuning &  9953\\ 
    Conversational Prompting & 1274\\
    \textbf{SimulMask} &  \textbf{1014} \\ 
    Causal Mask & 727 \\ 
    \hline
    \vspace{0.05in}
  \end{tabular}
  \caption{Time to complete one epoch for different fine-tuning approaches on an H100.}
  \label{tab:traintime}
\end{table}

\begin{figure}
    \centering
    \resizebox{0.5\textwidth}{!}{
        \begin{tikzpicture}
        \begin{axis}[
        every y tick label/.append style={font=\footnotesize},
        every x tick label/.append style={font=\footnotesize},
        yticklabel style={align=center},
        xlabel={Compuation (GFLOPs)},
        ytick={1,2,3,4},
        yticklabels={SM-norec-mod, converse-norec, prefix-rec, causal-rec},
        boxplot/variable width,
        boxplot/box extend=0.5,
        height=5cm,
        width=8cm,
        ]
        \addplot+ [
        boxplot prepared={
        lower whisker=49.87593459, lower quartile=112.9651628,
        median=160.3275265,
        upper quartile=228.827305, upper whisker=402.6205183},
        black, solid, thick] coordinates {};
        
        
        \addplot+ [
        boxplot prepared={
        lower whisker=55.14508821, lower quartile=131.4198449,
        median=189.3882094,
        upper quartile=273.873193, upper whisker=487.5532151},
        black, solid] coordinates {};

        \addplot+ [
        boxplot prepared={
        lower whisker=49.87593459, lower quartile=199.6511759,
        median=456.1252618,
        upper quartile=1095.620835, upper whisker=2439.575323},
        black, solid] coordinates {};
        
        \addplot+ [
        boxplot prepared={
        lower whisker=49.87593459, lower quartile=236.4628406,
        median=507.4421541,
        upper quartile=1168.551625, upper whisker=2566.684801},
        black, solid] coordinates {};
        
        \end{axis}
        \end{tikzpicture}
    }
    \caption{Box plots of the computational cost of each method in GFLOPs during inference.}
    \label{fig:computation}
\end{figure}
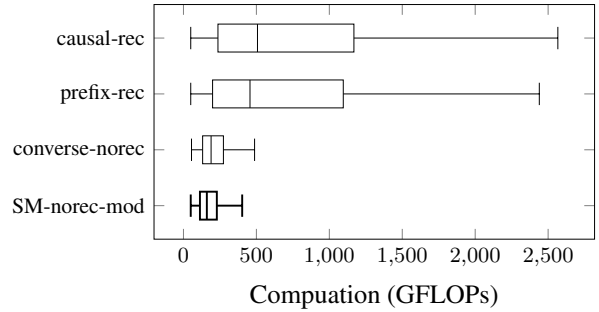

Furthermore, we find that \textit{SM-norec} is also more computationally efficient at inference than \textit{prefix-rec} and \textit{converse-norec}.  We report these results in GFLOPs that are needed to complete a sentence translation in Figure \ref{fig:computation}. The data used to obtain the results was a random 1000 samples from the English-French split of the IWSLT 2017 test set \cite{cettolo-etal-2017-overview}. The models chosen either used wait-3 or a chunk size of 5.

\begin{figure}
    \centering
    \resizebox{0.43\textwidth}{!}{
        \begin{tikzpicture}
        \begin{axis}[
            xlabel={\footnotesize Sequence Length},
            ylabel={\footnotesize Computation (GFLOPs)},
            every y tick label/.append style={font=\footnotesize},
            every x tick label/.append style={font=\footnotesize},
            yticklabel style={align=center},
            ymin=0, ymax=5500,
            xmin=0, xmax=140,
            height=4cm,
            width=6cm,
            stack plots=y,
            area style,
            legend style={
            at={(0.5,1.03)}, 
            anchor=south, 
            legend columns=-1 
            } 
        ]
        \addplot+[mark=none, fill=blue, fill opacity=0.5] coordinates {
            (0, 56.1813051955)
            (10,75.87703216125517)
            (20,101.17686380521741)
            (30, 126.89414690059333)
            (40, 153.34244311901668)
            (50, 178.69789411799005)
            (60, 204.71562304126317)
            (70, 231.1299038695536)
            (80, 258.05204357383786)
            (90, 283.28429528952375)
            (100, 310.1209130186333)
            (110, 337.2143020285)
            (120, 366.3721688469166)
            (130, 392.2513324452858)
            (140, 417.28598408683337)
        } \closedcycle;
        \addlegendentry{\footnotesize Initial} 
        
        \addplot+[mark=none, fill=red, fill opacity=0.5] coordinates {
            (0, 4.4646164019)
            (10, 36.32652225322069)
            (20, 94.25877662135404)
            (30, 197.18507117418)
            (40, 342.63493973419173)
            (50, 549.292903969198)
            (60, 780.4786706341315)
            (70, 1084.8430867923394)
            (80, 1427.211523196973)
            (90, 1846.2474545342618)
            (100, 2139.626294570267)
            (110, 2573.265723306889)
            (120, 3264.9883998733335)
            (130, 3681.7593976428575)
            (140, 4747.491182829666)
        } \closedcycle;
        \addlegendentry{\footnotesize Recompute} 
        
        
        \end{axis}
        \end{tikzpicture}
    }
    \caption{Separated computational cost in GFLOPs between initial (or required) computational cost and the cost of recomputing already emitted target words in a provided prompt during translation versus the sequence length of a given sample. }
    \label{fig:seqcomp}
\end{figure}
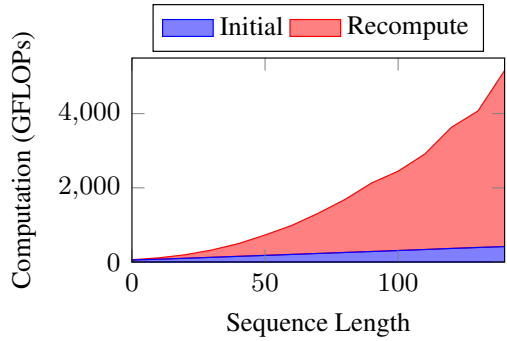

By leveraging SimulMask during fine-tuning, we eschew the need to recompute the KV cache at inference.  In doing so, SimulMask saves computation compared to \textit{prefix-rec} and \textit{causal-rec}.  We demonstrate the proportions of computation in GFLOPs dedicated to re-computing the KV cache and processing/predicting initial tokens in Figure \ref{fig:seqcomp} (based on \textit{prefix-rec}).  The sequence length is the number of tokens in the predicted target and input source. As can be seen, it is critical to avoid recomputing KV cache, as achieved by SimulMask, to provide low latency translations, especially at longer sequence lengths.  

\section{Conclusion}
In this work, we first examine current LLM fine-tuning approaches for SimulMT and identify their shortcomings. We then propose a new paradigm for fine-tuning LLMs for SimulMT that we call SimulMask, which avoids the shortcomings of previous methods.  When employing SimulMask, the target sequence is prevented from attending to a portion of the source sequence according to an arbitrary decision policy modeling SimulMT.  Through the application of SimulMask, we can efficiently fine-tune a LLM for SimulMT and reduce the computational costs of inference by eliminating the recomputation of the KV cache for the target sequence, unlike prior works.  Furthermore, we can exceed or match the translation quality of prior works at all wait-k values across multiple language pairs.

\section*{Limitations}
Given the translation quality benefits at a reduced computational cost of fine-tuning with SimulMask, it would be beneficial to evaluate the approach to larger and more powerful LLMs, adapting them for SimulMT. Also, while SimulMask is broadly applicable to various decision policies, our current evaluation was limited to only testing the effectiveness of SimulMask on the wait-k policy and did not evaluate alternative fixed or more flexible decision policies.  Additionally, we did not explore simultaneous speech-to-text or speech-to-speech translation, which SimulMask has yet to be tested on.

\section*{Acknowledgments}
This research was supported, in part, by the National Science Foundation grants 2223483 and 2223484.

\bibliography{custom}


\appendix
\section{Appendix}
\subsection{Encoder Attention Mask}
\label{app:enc}
\begin{figure}[h!]
    \centering
    \includegraphics[width=0.175\textwidth]{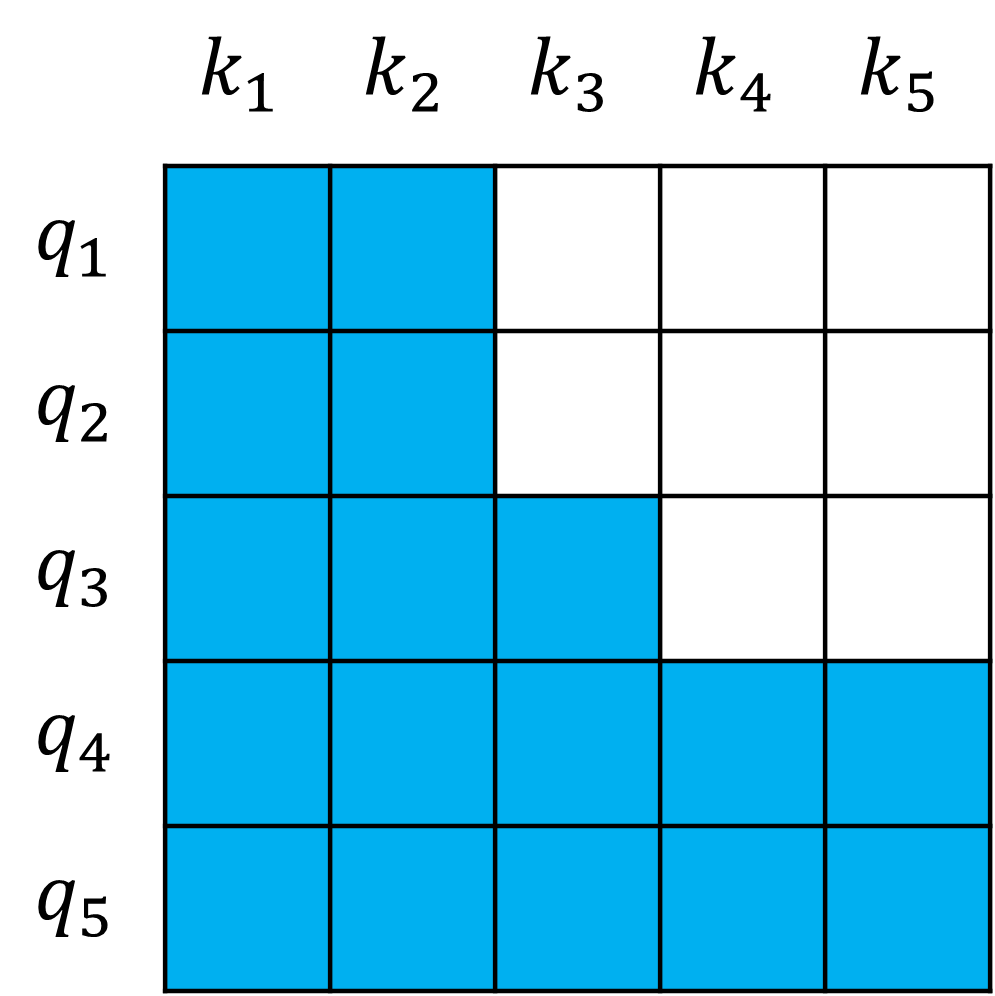}
    \caption{An attention mask to model simultaneous translation for a Transformer encoder during training.}
    \label{fig:encmask}
\end{figure}

An example of an encoder attention mask used to model simultaneous translation during training is provided in Figure \ref{fig:encmask}.  The attention mask in Figure \ref{fig:encmask} is designed for a source sequence length of 5 tokens where the first read step reads 2 tokens, the second read step reads 1 token, and the third read step reads 2 tokens.

\subsection{Inference Mirrored Attention and SimulMask}
\label{app:align}

    \begin{figure*}[h!]
    \centering
    \begin{subfigure}[h]{\columnwidth}
        \centering
        \includegraphics[width=0.35\linewidth]{Figures/InferenceAlignments.png}
    \caption{Attention for the first prediction step.}
    \label{fig:appinference1}
    \end{subfigure}\hfill
    \begin{subfigure}[h]{\columnwidth}
           \centering 
          \includegraphics[width=0.9\linewidth]{Figures/PartialAttention.png}
        \caption{Inference mirrored attention for the first prediction step.}
        \label{fig:appfinetune1}
    \end{subfigure}
    \begin{subfigure}[h]{\columnwidth}
        \centering
        \includegraphics[width=0.5\linewidth]{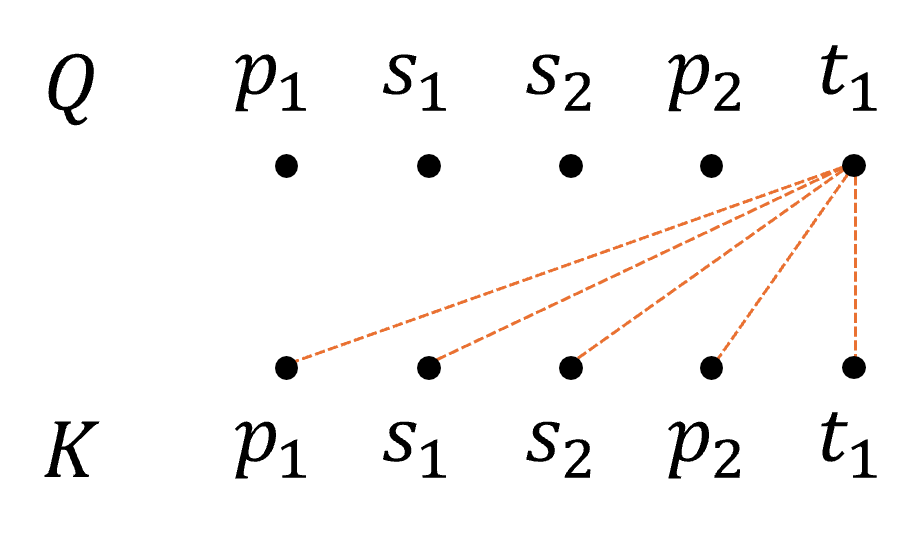}
    \caption{Attention for the second prediction step.}
    \label{fig:appinference2}
    \end{subfigure}\hfill
    \begin{subfigure}[t]{\columnwidth}
           \centering 
          \includegraphics[width=0.9\linewidth]{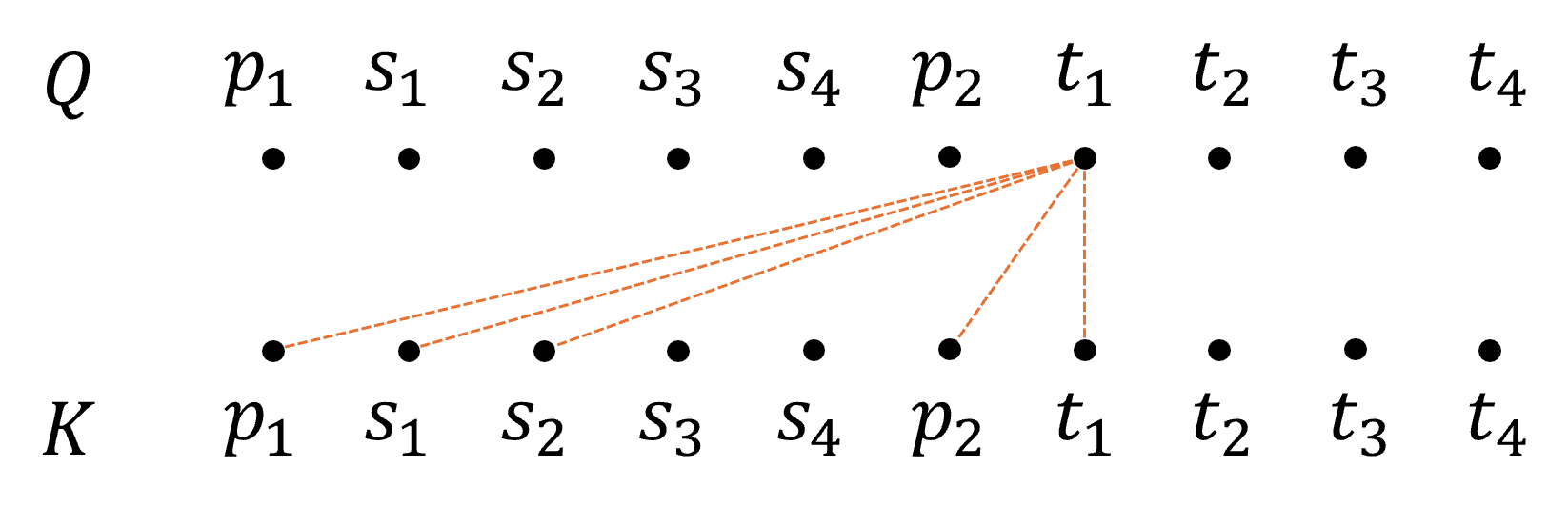}
        \caption{Inference mirrored attention for the second prediction step.}
        \label{fig:appfinetune2}
    \end{subfigure}
    \begin{subfigure}[h]{\columnwidth}
        \centering
        \includegraphics[width=0.7\linewidth]{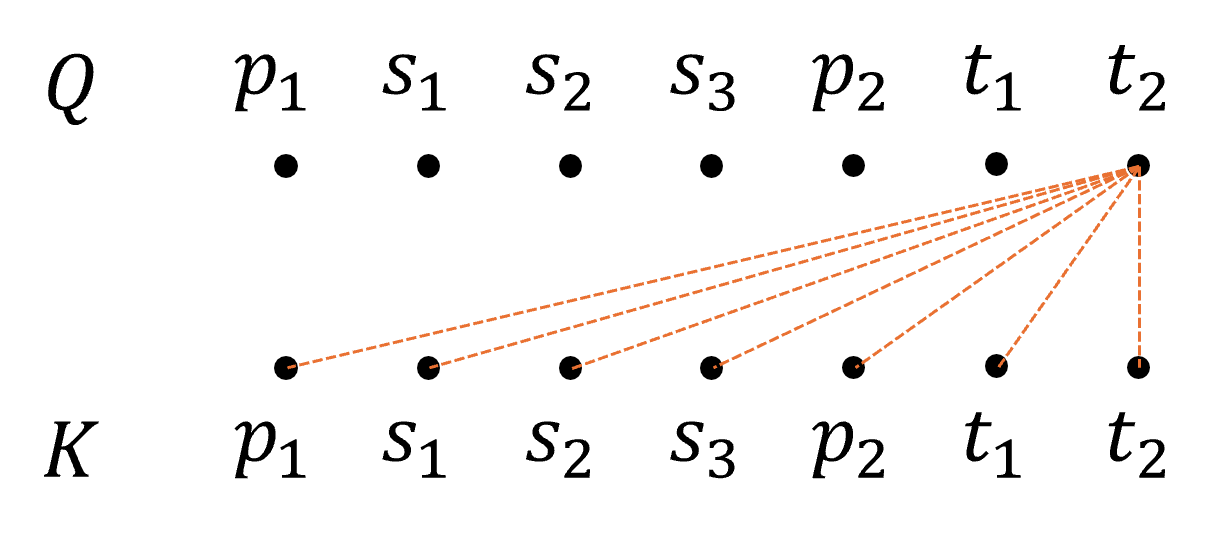}
    \caption{Attention for the third prediction step.}
    \label{fig:appinference3}
    \end{subfigure}\hfill
    \begin{subfigure}[h]{\columnwidth}
           \centering 
          \includegraphics[width=0.9\linewidth]{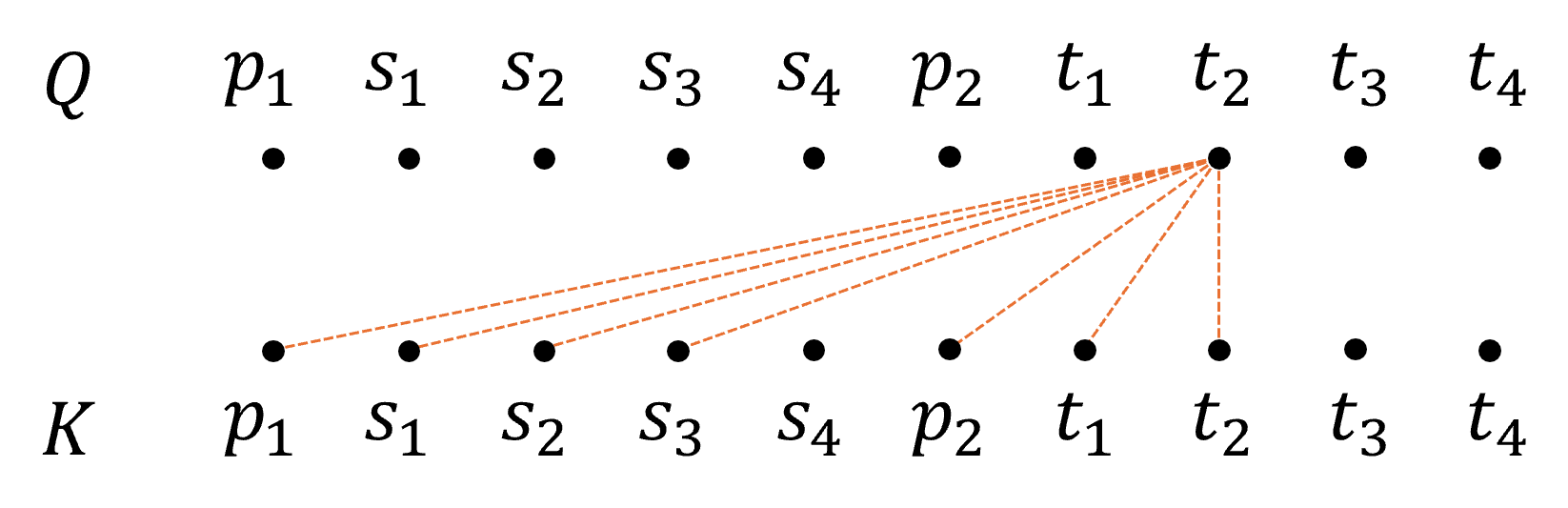}
        \caption{Inference mirrored attention for the third prediction step.}
        \label{fig:appfinetune3}
    \end{subfigure}
\caption{Attention during inference and finetuning for SimulMT.}
\label{fig:attentionalignment}
\end{figure*}

Figure \ref{fig:attentionalignment} provides the complete example of inference mirrored attention for the wait-1 policy explained in Section \ref{Sec:restrictattention}.  To reiterate in Figure \ref{fig:appinference1}, the query of $p_2$ attends to the keys of $p_1, s_1, p_2$.  Thus, during fine-tuning, to eliminate the fine-tuning-inference mismatch, the query of $p_2$ is limited to similarly attend to the keys of $p_1, s_1, p_2$ as shown in Figure \ref{fig:appfinetune1} rather than the entire source sequence.  Then for the second and third decoding steps at inference, the queries of $t_1$ and $t_2$ attend to the keys of $p_1, s_1, s_2, p_2$ and $p_1, s_1, s_2, s_3, p_2$, respectively as shown in Figures \ref{fig:appinference2} and \ref{fig:appinference3}.  Once again, to eliminate the fine-tuning-inference mismatch, the queries of $t_1$ and $t_2$ must attend to an identical set of keys as shown in Figures \ref{fig:appfinetune2} and \ref{fig:appfinetune3}.

\subsection{Hyperparameters}
\label{app:hyperparams}
\begin{table*}[h!]
    \centering
    \begin{tabular}{ccccc}
    \hline
     \textbf{Hyperparameter} & \textbf{Group 1} & \textbf{Group 2} & 
     \textbf{Group 3} &\textbf{Group 4} \\ 
     \hline
     \textbf{Weight Precision} & bfloat16  & bfloat16 & bfloat16 & bfloat16 \\
    \textbf{Optimizer} & AdamW & AdamW & AdamW & AdamW \\
    \textbf{Learning Rate} & $2\cdot10^{-4}$ & $2\cdot10^{-4}$ & $2\cdot10^{-4}$ & $2\cdot10^{-4}$ \\
    \textbf{LR Scheduler} & Inverse Sqrt & Inverse Sqrt & Inverse Sqrt & Inverse Sqrt \\
    \textbf{Weight Decay} & 0.1 & 0.1 & 0.1 & 0.1 \\
    \textbf{Warmup Ratio} & 0.03 & 0.03 & 0.03 & 0.03 \\
    \textbf{Max Gradient Norm} & 1 & 1 & 1 & 1 \\
    \textbf{Max Sequence Length} & 512 & 512 & 512 & 512 \\
    \textbf{Wait-k} & 5,7,9,11 & 5,7,9,11 & - & - \\
    \textbf{Epochs} & 2 & 1 & 1 & 2 \\
    \textbf{Batch size} & 64 & 1024 & 64 & 64 \\
    \textbf{Attention heads} & 32  & 32 & 32 & 32 \\
    \textbf{Layers} & 24 & 24 & 24 & 24 \\
    \textbf{Hidden Size} & 2048 & 2048 & 2048 & 2048 \\
    \textbf{Positional Encoding} & Modified ALiBi & ALiBi & ALiBi & ALiBi \\
    \textbf{Attention Mask} & SimulMask & Causal Mask & Causal Mask & Causal Mask \\
    $\mathbf{\delta_{max}}$ & - & - & 10 & - \\
    $\mathbf{\beta}$ & - & - & 0.5 & - \\
    $\mathbf{\rho_{min}}$ & - & - & 0.5 & - \\
    $\mathbf{\rho_{max}}$ & - & - & 0.9 & - \\
    \hline
    \end{tabular}
    \caption{Fine-tuning hyperparameters for all models in Section \ref{Sec:Results}. \textbf{Group 1:} \textit{SM-norec, SM-norec-mod}. \textbf{Group 2:} \textit{prefix-rec, prefix-norec}. \textbf{Group 3:} converse-norec. \textbf{Group 4:} \textit{causal-rec, causal-offline}.}
    \label{tab:all_hyperparams}
\end{table*}

The fine-tuning hyperparameters used for \textit{SM-norec, causal-rec, causal-offline,  prefix-rec, prefix-norec,} and \textit{converse-norec} models are provided in Table \ref{tab:all_hyperparams}. \\ 

The prompts used for the \textit{SM-norec, causal-rec, causal-offline, prefix-rec,} and \textit{prefix-norec} models consisted of the following format: 

\begin{verbatim}
    Translate the following sentence from 
    [SRC] to [TGT]: [SRC-Sentence]\n
    Assistant: [TGT-Sentence]
\end{verbatim}

Alternatively, the \textit{converse-norec} model used the prompt: 

\begin{verbatim}
Translate the following sentence from 
[SRC] to [TGT]\nAssistant: <s><t>
[SRC-Chunk-1]</t>[TGT-Chunk-1]
</s><s><t>...<s><t>
[SRC-Chunk-n]</t>[TGT-Chunk-n]</s>
\end{verbatim}

Our implementation for \textit{converse-norec} followed \citet{wang2024conversational}.  However, we used the Itermax method from the SimAlign toolkit leveraging XLM-RoBERTa base \cite{conneau2019unsupervised} to align words due to their work reporting better alignments than fast-align \cite{jalili-sabet-etal-2020-simalign, dyer-etal-2013-simple}.

\subsection{Extended Translation Results}
\label{app:transresults}

\begin{table*}[h!]
    \centering
    \begin{tabular}{@{\hspace{1pt}}l|ccccc@{\hspace{1pt}}}
    \hline
     \textbf{Model} & \textbf{en-fr} & \textbf{en-nl} & \textbf{en-it} &\textbf{en-ro} & \textbf{en-de }\\ 
     \hline
     SM-norec-mod (wait-1) & 28.89 (2.05)  & 16.37 (1.53) & 15.72 (1.92) & 9.77 (1.53) & 17.73 (1.68) \\
     SM-norec-mod (wait-3) & 36.48 (3.67)  & 25.31 (3.37) & 23.17 (3.29) & 21.44 (3.32) & 24.34 (3.23) \\
     SM-norec-mod (wait-5) & 38.77 (5.39)  & 29.13 (5.13) & 28.01 (5.09) & 25.79 (5.18) & 27.11 (4.92) \\
     SM-norec-mod (wait-7) & 38.85 (7.02)  & 29.98 (6.78) & 29.81 (6.76) & 26.07 (6.82) & 28.43 (6.58) \\
     \hline
     prefix-rec (wait-1) & 16.37 (1.64) & 4.53 (1.14) & 5.27 (1.12) & 4.03 (1.29) & 5.64 (2.12)\\
     prefix-rec (wait-3) & 28.90 (3.47) & 22.44 (3.25) & 21.74 (3.15) & 14.82 (3.29) & 19.13 (3.09)  \\
     prefix-rec (wait-5) & 39.55 (5.33) & 29.76 (5.17) & 27.26 (5.03) & 20.85 (5.14) & 27.38 (4.86) \\
     prefix-rec (wait-7) & 40.48 (7.00) & 30.95 (6.77) & 30.52 (6.75) & 25.06 (6.78) & 28.79 (6.56) \\
     \hline
     converse-norec (chunk-1) & 19.89 (0.95) & 9.17 (0.65) & 10.86 (0.89) & 10.80 (0.82) & 13.67 (1.13) \\
     converse-norec (chunk-3) & 31.94 (2.46) & 24.22 (2.76) & 23.04 (2.98) & 21.02 (2.86) & 24.26 (2.83) \\
     converse-norec (chunk-5) & 35.03 (3.55) & 25.49 (3.99) & 24.82 (4.15) & 22.89 (4.05) & 26.17 (4.04) \\
     converse-norec (chunk-7) & 36.42 (4.67) & 27.09 (5.17) & 27.40 (5.32) & 23.48 (5.18) & 26.82 (5.22) \\
     converse-norec (chunk-9) & 36.87 (5.79) & 27.86 (6.29) & 27.37 (6.37) & 23.73 (6.31) & 27.28 (6.39) \\
     converse-norec (chunk-11) & 37.27 (6.87) & 27.84 (7.37) & 27.52 (7.42) & 23.76 (7.40) & 27.47 (7.47) \\
    \hline
    \end{tabular}
    \caption{Translation quality and latency results in BLEU and LAAL.}
    \label{tab:BLEU}
\end{table*}

\begin{table*}[h!]
    \centering
    \begin{tabular}{@{\hspace{1pt}}l|ccccc@{\hspace{1pt}}}
    \hline
     \textbf{Model} & \textbf{en-fr} & \textbf{en-nl} & \textbf{en-it} & \textbf{en-ro} & \textbf{en-de} \\ 
     \hline
     SM-norec-mod (wait-1) & 50.76 (2.05) & 36.14 (1.53) & 36.37 (1.92) &  26.36 (1.53) & 42.20 (1.68)\\
     SM-norec-mod (wait-3) & 58.42 (3.67) & 47.60 (3.37) & 45.17 (3.29) & 43.41 (3.32) & 50.64 (3.23)\\
     SM-norec-mod (wait-5) & 60.40 (5.39) & 52.34 (5.13) & 51.34 (5.09) & 49.78 (5.18) & 53.09 (4.92)\\
     SM-norec-mod (wait-7) & 60.59 (7.02) & 53.74 (6.78) & 53.30 (6.76) & 50.18 (6.82) & 53.73 (6.58)\\
     \hline
     prefix-rec (wait-1) & 33.72 (1.64) & 15.50 (1.14) & 18.72 (1.12) & 13.62 (1.18) & 25.71 (2.16) \\
     prefix-rec (wait-3) & 49.64 (3.47) & 44.26 (3.25) & 42.63 (3.15) & 28.24 (3.173) & 40.15 (3.10) \\
     prefix-rec (wait-5) & 61.05 (5.33) & 53.50 (5.17) & 49.91 (5.03) & 43.11 (5.04) & 52.36 (4.86) \\
     prefix-rec (wait-7) & 61.91 (7.00) & 55.07 (6.77) & 54.28 (6.75) & 49.09 (6.77) & 53.46 (6.53) \\
     \hline
     converse-norec (chunk-1) & 47.54 (0.95) & 28.01 (0.65) & 33.19 (0.89)  & 30.67 (0.74) & 46.12 (1.24) \\
     converse-norec (chunk-3) & 56.45 (2.46) & 47.13 (2.76) & 46.93 (2.98)  & 44.04 (2.74) & 52.35 (2.91) \\
     converse-norec (chunk-5) & 58.27 (3.55) & 49.15 (3.99) & 48.98 (4.15)  & 46.49 (3.96) & 53.05 (4.09) \\
     converse-norec (chunk-7) & 58.98 (4.67) & 50.69 (5.17) & 51.37 (5.32) & 47.97 (5.08) & 53.11 (5.26) \\
     converse-norec (chunk-9) & 59.33 (5.79) & 51.52 (6.29) & 51.49 (6.37) & 48.56 (6.22) & 53.14 (6.38) \\
     converse-norec (chunk-11) & 59.47 (6.87) & 52.11 (7.37) & 51.47 (7.42) & 48.85 (7.31) & 52.97 (7.43) \\
    \hline
    \end{tabular}
    \caption[Caption for LOF]{Translation quality and latency results in chrF++ and LAAL.\footnotemark}
    \label{tab:CHRF}
\end{table*}

The translation quality results in Table \ref{tab:BLEU} provide the numerical BLEU and LAAL values from Figure \ref{fig:translationresults} for \textit{prefix-rec}, \textit{converse-norec}, and \textit{SM-norec-mod}. Alternatively, Table \ref{tab:CHRF} provides the numerical chrF++ values associated with each of these models from Figure \ref{fig:translationresults}.\\

\subsection{Sequence Lengths}
\label{app:compext}
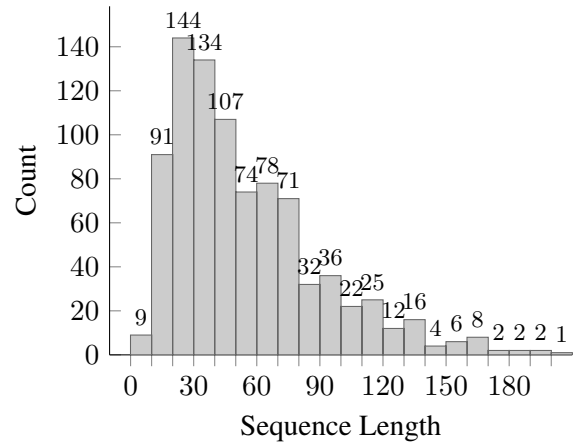
\begin{figure}[h]
        \centering
        \begin{tikzpicture}
            \begin{axis}[
                ybar = 0.5,
                height=0.25\textheight,
                width=\columnwidth,
                ymin=0,
                xmin=0, xmax=200,
                enlarge x limits=0.05,
                xticklabels={$0$, , ,$30$, , ,$60$, , ,$90$, , ,$120$, , ,$150$, , ,$180$, , ,$210$},
                bar width=8pt,
                bar shift=5,
                ylabel={Count},
                xlabel={Sequence Length},
                nodes near coords,
                every node near coord/.append style={font=\footnotesize},
                axis lines*=left,
                yticklabels={$0$, $20$, $40$, $60$, $80$, $100$, $120$, $140$, $160$},
                ytick={0, 20, 40, 60, 80, 100, 120, 140, 160},
                xtick=data
            ]
                \addplot[fill=black!20,draw=black!60] coordinates {        
                (0, 9) 
                (10, 91) 
                (20, 144) 
                (30, 134) 
                (40, 107) 
                (50, 74) 
                (60, 78)
                (70, 71)
                (80, 32)
                (90, 36)
                (100, 22)
                (110, 25)
                (120, 12)
                (130, 16)
                (140, 4)
                (150, 6)
                (160, 8)
                (170, 2)
                (180, 2)
                (190, 2)
                (200, 1)
                };
            \end{axis}
        \end{tikzpicture}
        \caption{Histogram of the distribution of sequence lengths.}
        \label{fig:histo}
    \end{figure}

Figure \ref{fig:histo} reports the number of occurrences on the English-French IWSLT2017 validation set \cite{cettolo-etal-2017-overview} that the combined length of the source sequence and the predicted target sequence are within a specified range for prefix-rec at wait-3. 

\subsection{Licensing Information}
The SimulEval toolkit is licensed under CC BY-SA-4.0 license \cite{ma-etal-2020-simuleval}.  The Simul-LLM framework and SimAlign toolkit are licensed under the MIT license \cite{agostinelli2023simul, jalili-sabet-etal-2020-simalign}.  The IWSLT 2017 dataset is licensed under CC BY-NC-ND \cite{cettolo-etal-2017-overview}.  The Falcon model we used from Hugging Face (tiiuae/falcon-rw-1b) is licensed under Apache 2.0 \cite{refinedweb}.
     \footnotetext{The chrF++ results for \textit{prefix-rec} and \textit{converse-norec} on the English-German and English-Romanian language pairs were obtained with different fine-tuning and evaluation seeds than the models reported in Figure \ref{fig:translationresults}.}

\end{document}